\documentclass[conference]{IEEEtran}
\IEEEoverridecommandlockouts
\usepackage{cite}
\usepackage{amsmath,amssymb,amsfonts}
\usepackage{algorithmic}
\usepackage{graphicx}
\usepackage{textcomp}
\usepackage{xcolor}
\usepackage{algorithm}
\usepackage{booktabs}
\usepackage{multirow}
\usepackage{subcaption}
\def\BibTeX{{\rm B\kern-.05em{\sc i\kern-.025em b}\kern-.08em
    T\kern-.1667em\lower.7ex\hbox{E}\kern-.125emX}}
\begin{document}

\title{SLA-MORL: SLA-Aware Multi-Objective Reinforcement Learning for HPC Resource Optimization}


\author{\IEEEauthorblockN{Seraj Al Mahmud Mostafa$^{1}$, Aravind Mohan$^{2}$, Jianwu Wang$^{1}$}
\IEEEauthorblockA{\textit{$^1$Department of Information Systems, University of
Maryland, Baltimore County,} Baltimore, MD, USA\\
\textit{$^2$Department of Computer Science, McMurry University,} Abilene, TX, USA\\
}
}

\maketitle

\begin{abstract}
Dynamic resource allocation for machine learning workloads in cloud environments remains challenging due to competing objectives of minimizing training time and operational costs while meeting Service Level Agreement (SLA) constraints. Traditional approaches employ static resource allocation or single-objective optimization, leading to either SLA violations or resource waste. We present SLA-MORL, an adaptive multi-objective reinforcement learning framework that intelligently allocates GPU and CPU resources based on user-defined preferences (time, cost, or balanced) while ensuring SLA compliance. Our approach introduces two key innovations: (1) intelligent initialization through historical learning or efficient baseline runs that eliminates cold-start problems, reducing initial exploration overhead by 60\%, and (2) dynamic weight adaptation that automatically adjusts optimization priorities based on real-time SLA violation severity, creating a self-correcting system. SLA-MORL constructs a 21-dimensional state representation capturing resource utilization, training progress, and SLA compliance, enabling an actor-critic network to make informed allocation decisions across 9 possible actions. Extensive evaluation on 13 diverse ML workloads using production HPC infrastructure demonstrates that SLA-MORL achieves 67.2\% reduction in training time for deadline-critical jobs, 68.8\% reduction in costs for budget-constrained workloads, and 73.4\% improvement in overall SLA compliance compared to static baselines. By addressing both cold-start inefficiency and dynamic adaptation challenges, SLA-MORL provides a practical solution for cloud resource management that balances performance, cost, and reliability in modern ML training environments. 
\end{abstract}

\section{Introduction}
\label{sec:introduction}

Modern High-Performance Computing (HPC) and Cloud environments increasingly support complex AI models that process datasets varying widely in size and complexity, necessitating careful alignment between available resources such as CPUs and GPUs, and workload requirements. Traditional management strategies struggle with dynamically matching resource allocation to these diverse and fluctuating demands, often leading to inefficient utilization, increased operational costs, and compromised performance, while over-provisioning expensive hardware drives operational costs skyward with marginal performance gains. Multi-Objective Reinforcement Learning (MORL), a RL mechanism \cite{sutton1998reinforcement} to tackle various objectives simultaneously, emerges as a promising solution by learning to adaptively optimize resources through real-time interactions while balancing multiple competing Service Level Agreement (SLA) objectives such as cost efficiency, training time, and resource utilization.

However, applying MORL to complex HPC and Cloud environments introduces significant challenges. First, RL algorithms suffer from sample inefficiency \cite{ball2023efficient}, requiring extensive interactions with dynamic systems, thus incurring high computational costs and prolonged training periods \cite{gu2025deep}. Second, traditional RL methods fail to rapidly adapt to workload pattern changes and demand spikes, leading to costly overprovisioning or detrimental SLA violations \cite{jimenez2023resource}. Third, multi-objective optimization compounds these issues as RL policies typically prioritize single metrics, neglecting crucial Quality of Service (QoS) tradeoffs and resulting in resource hoarding \cite{mao2022mean}. Furthermore, RL agents lack built-in monitoring for resource misuse while inadequate state representations and limited resource capability knowledge lead to suboptimal configurations \cite{gu2025deep}. Finally, algorithmic instability, poor sim-to-real generalization, and exploration-exploitation imbalances further hinder deployment, risking frequent retraining and eroded user trust \cite{hortelano2023comprehensive}.

Current resource optimization approaches suffer from either static allocation strategies that ignore dynamic workload patterns, or reactive adaptive methods that lack workload-awareness and SLA coordination, resulting in inefficient resource utilization, increased operational costs, and compromised performance. We identify two critical research challenges: \textbf{(1)} \textit{Cold-Start and Exploration Inefficiency in Existing Methods}: Traditional approaches begin with random resource allocations and lack mechanisms to leverage historical patterns, forcing costly exploration from scratch for every new workload; \textbf{(2)} \textit{Static Multi-Objective Optimization}: Current systems employ fixed weight parameters for competing objectives, failing to adapt dynamically to SLA violations and changing system states.

To tackle these challenges, we introduce \textbf{SLA-MORL}, a multi-objective reinforcement learning framework that accepts model-data pairs and user-defined SLA preferences as input, and produces optimized resource allocation solutions with comprehensive performance analytics. The framework addresses dynamic resource allocation through two key innovations: \textbf{(1)} \textit{Intelligent Initialization with Optional Historical Learning}: SLA-MORL eliminates cold-start problems through an adaptive actor-critic architecture that either pre-trains from historical logs when available or employs efficient baseline runs (10\% epochs, 20\% data) for new workloads, ensuring intelligent starting configurations without costly random exploration; \textbf{(2)} \textit{SLA-aware Pareto Optimization}: Our novel contribution unifies SLA violation detection and adaptive weight computation into a dynamic reward system that automatically adjusts optimization priorities based on violation severity.

SLA-MORL supports three user preferences: \textit{time} (minimize training time using maximum resources where cost can increase), \textit{cost} (minimize expense while allowing extended training time), and \textit{balanced} (optimize both objectives with optional user-specified targets). This adaptive capability is formalized through our core optimization objective:

\begin{equation}\label{eq:action}
a_t = \arg\max_{a \in \mathcal{A}} \sum_{i \in \{time, cost, util\}} w_i^{(t)} R_i(s_t, a),
\end{equation}

\noindent where $w_i^{(t)}$ represent adaptive weights that evolve according to SLA violation patterns and user preferences, $s_t$ captures workload characteristics and system state enhanced with preference encoding, and $a \in \mathcal{A}$ represents resource configurations (GPU, CPU).

We evaluate SLA-MORL on 13 diverse ML workloads using production HPC infrastructure with NVIDIA RTX 8000 GPUs. Our experiments demonstrate significant improvements over static baselines: 67.2\% reduction in training time, 68.8\% reduction in operational costs, and 73.4\% improvement in SLA compliance. These results validate that our two technical innovations (historical learning for intelligent initialization and adaptive weight computation for SLA aware optimization) successfully address the fundamental challenges in cloud resource allocation for ML workloads.

\section{Related Work}
\label{sec:relatedwork}

\noindent\textbf{Resource Optimization with Reinforcement Learning.} Recent advances in deep reinforcement learning have shown promise for dynamic resource allocation in cloud environments. Chen et al.~\cite{chen2022adaptive} propose an A3C-based resource allocation method that combines actor-critic networks to optimize QoS and energy efficiency. While their approach demonstrates superior performance over traditional heuristics, it employs static weight parameters that cannot adapt to SLA violations. Similarly, Zhao et al.~\cite{zhao2022performance} present a DDPG-based task scheduling framework with correlation-aware state representations. However, both approaches optimize for fixed objectives without considering dynamic SLA requirements.

Liu et al.~\cite{liu2017hierarchical} advance the field with a hierarchical DRL framework combining global VM placement with LSTM-powered local server control, achieving significant energy and latency reductions. Baheri et al.~\cite{baheri2022mars} introduce MARS, an SLA-aware scheduler that combines offline heuristics with runtime A3C optimization. While MARS considers SLA constraints, it lacks our adaptive weight mechanism and suffers from cold-start problems when historical data is unavailable.

\noindent\textbf{Offline to Online Learning Approaches.} The integration of offline and online learning has emerged as a solution for improving sample efficiency in RL systems. Zheng et al.~\cite{zheng2023adaptive} introduce Adaptive Policy Learning (APL), where agents pre-train on static datasets before adapting with online feedback. Wang et al.~\cite{wang2023train} propose FamO2O, which pre-trains multiple policies and selects them adaptively per state. Ball et al.~\cite{ball2023efficient} present RLPD, accelerating online learning through symmetric sampling and ensemble methods.

These offline-to-online approaches primarily target robotics domains (Niu et al.~\cite{niu2022trust}) or multi-agent environments (Liu et al.~\cite{liu2025offline}). None address the unique challenges of cloud resource optimization where workloads vary dramatically and SLA violations carry immediate financial consequences. Our work uniquely combines historical learning with dynamic SLA-aware adaptation for cloud environments.

\noindent\textbf{Multi-Objective Reinforcement Learning.} Multi-objective optimization in RL has received significant attention for balancing conflicting goals. Nguyen et al.~\cite{nguyen2021prioritized} propose two actor-critic methods (MCSP and SCMP) that integrate scalarized rewards into policy optimization. Liu et al.~\cite{liu2025pareto} introduce PSL-MORL using hypernetworks to generate personalized policies across the Pareto front. Cai et al.~\cite{cai2023distributional} extend Pareto-optimal MORL with distributional approaches using utility-based reward shaping, Rakshit et al.~\cite{rakshit2024righteous} introduces informed Pareto Simulated Annealing (iPSA) to efficiently explore the trade-offs in deployment configurations.

While these MORL methods advance the theoretical foundations, they operate primarily in simulated environments with static weight parameters. Zhou et al.~\cite{zhou2023gradient} make progress with gradient-adaptive Pareto optimization for constrained RL, but their approach lacks integration with real-world SLA constraints. Qin et al.~\cite{qin2021multi} address deadline-constrained workflow scheduling through DCMORL, yet still rely on fixed Chebyshev scalarization without adaptive mechanisms.

\noindent\textbf{SLA-Aware Resource Management.} SLA compliance represents a critical requirement in production cloud systems. Souza et al.~\cite{souza2021hpc} propose ASAX, an RL-based HPC co-scheduler using decision tree experts to maintain QoS constraints, improving utilization by up to 51\%. Haratian et al.~\cite{haratian2017adaptive} present AFRM, a fuzzy resource management framework that reduces SLA violations through dynamic policies. Dai et al.~\cite{dai2023augmented} enhance constraint satisfaction through augmented proximal policy optimization.

However, these SLA-aware approaches share common limitations: they lack learning from historical patterns, operate with discrete action spaces (Ran et al.~\cite{ran2019slas}), or focus on single objectives (Peng et al.~\cite{peng2016reinforcement}). None provide the adaptive weight mechanisms necessary for handling dynamic SLA violations in multi-objective settings.

\noindent\textbf{Research Gap and Our Contribution.} Existing approaches fall into three categories, each with critical limitations:

\textbf{Single-objective RL methods} (Chen et al.~\cite{chen2022adaptive}, Baheri et al.~\cite{baheri2022mars}) use static weights and cannot adapt when SLA violations occur. \textbf{General MORL frameworks} (Nguyen et al.~\cite{nguyen2021prioritized}, Liu et al.~\cite{liu2025pareto}) lack real-world cloud integration and operate in simulated environments. \textbf{SLA-aware systems} (Souza et al.~\cite{souza2021hpc}, Dai et al.~\cite{dai2023augmented}) miss opportunities for historical learning and adaptive optimization.

SLA-MORL uniquely addresses these gaps through: (1) intelligent initialization that leverages historical patterns or efficient baseline runs to eliminate cold-start problems, and (2) dynamic weight adaptation based on SLA violation severity, enabling self-correcting resource allocation. Unlike existing work that treats these challenges separately, we provide an integrated solution specifically designed for production cloud environments where both rapid adaptation and SLA compliance are critical.

\begin{figure}[htbp!]
\centering
\includegraphics[width=\linewidth]{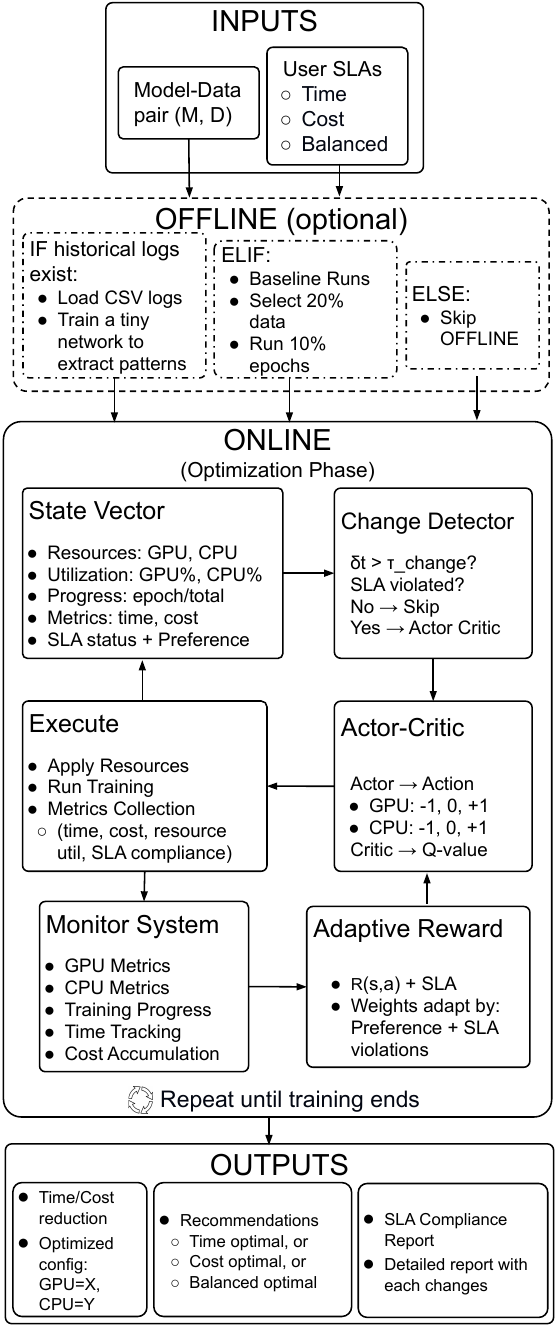} 
\caption{Proposed SLA-MORL Architecture.}
\label{fig:arch}
\end{figure}

\section{Methodology}
\label{sec:methodology}

We propose SLA-MORL, a dynamic resource allocation framework for ML training workloads that intelligently aligns computational resources (GPU, CPU) with user-defined SLA preferences through adaptive multi-objective reinforcement learning.

\subsection{System Overview}

Figure~\ref{fig:arch} illustrates the SLA-MORL architecture, and the complete process follows Algorithm~\ref{alg:SLA-MORL-overall}, which operates in four phases:

\noindent\textbf{Input Phase.} The system accepts two inputs: (1) a model-data pair consisting of ML training code and dataset, and (2) user-defined SLA preferences specified as either \texttt{time} priority (e.g., complete within 60 minutes), \texttt{cost} priority (e.g., spend less than \$20), or \texttt{balanced} (e.g., optimize both within given targets).

\noindent\textbf{Offline Phase (Optional).} Before optimization begins, SLA-MORL can leverage prior knowledge through three possible paths: (1) load historical CSV logs and extract patterns to pre-train the Actor-Critic networks, (2) run baseline experiments using 20\% of data for 10\% of epochs to gather initial performance estimates, or (3) skip this phase entirely and proceed directly to online optimization.

\noindent\textbf{Online Phase.} The core optimization loop executes continuously throughout model training with five interacting components. The \textbf{State Vector} constructs a 21-dimensional representation of system status. The \textbf{Change Detector} monitors for significant changes or SLA violations. The \textbf{Actor-Critic Network} makes resource allocation decisions. The \textbf{Execute} function implements resource changes through system APIs. The \textbf{Adaptive Reward} function computes feedback signals that dynamically adjust based on SLA compliance, forming a closed-loop system.

\noindent\textbf{Output Phase.} The system produces comprehensive reports including final training metrics, optimal resource configurations, SLA compliance status, and recommendations.

\begin{algorithm}[htbp]
\caption{SLA-MORL Overall Training Process}
\label{alg:SLA-MORL-overall}
\begin{algorithmic}[1]
\STATE \textbf{Input:} Model-data pair $(M, D)$, user preference $p \in \{time, cost, balanced\}$
\STATE \textbf{Output:} Optimized resource allocation $(g^*, c^*)$ and performance report
\STATE Initialize system parameters: $\tau_{change} = 0.1$, $\alpha = 0.5$, $\gamma = 0.95$
\STATE Execute Phase 1: Historical Learning or Baseline Initialization (Algorithm~\ref{alg:phase1})
\STATE Execute Phase 2: Online Optimization with trained/initialized policy (Algorithm~\ref{alg:phase2})
\STATE Generate Pareto front from execution history
\STATE Select best configuration $(g^*, c^*)$ based on user preference $p$
\STATE Return performance report with SLA compliance analysis
\end{algorithmic}
\end{algorithm}

\subsection{MDP Formulation for SLA-Aware Resource Allocation}

We formalize the dynamic resource allocation problem as a Markov Decision Process (MDP) defined by the tuple $(\mathcal{S}, \mathcal{A}, \mathcal{T}, \mathcal{R}, \gamma)$.

\subsubsection{State Space Design}

The state vector $\mathbf{s}_t \in \mathbb{R}^{21}$ captures system dynamics through six components:

\begin{equation}\label{eq:state-vector}
\mathbf{s}_t = [\mathbf{r}_t, \mathbf{u}_t, \mathbf{p}_t, \mathbf{c}_t, \mathbf{v}_t, \mathbf{w}_t],
\end{equation}

\noindent where \textbf{Resource Allocation} $\mathbf{r}_t = [g_t, c_t]$ represents current GPU count and CPU cores; \textbf{Resource Utilization} $\mathbf{u}_t = [u^{gpu}_t, u^{cpu}_t]$ indicates normalized usage rates; \textbf{Progress Indicators} $\mathbf{p}_t = [p^{epoch}_t, p^{throughput}_t]$ tracks training progress; \textbf{SLA Compliance Flags} $\mathbf{c}_t \in \{0,1\}^6$ monitors constraint satisfaction for each preference type; \textbf{Violation Severity} $\mathbf{v}_t \in [0,1]^6$ quantifies violation magnitude; and \textbf{User Preference} $\mathbf{w}_t \in \{0,1\}^3$ encodes the selected priority mode.

\subsubsection{Action Space and Resource Models}

We define a discrete action space for stable resource control:

\begin{equation}\label{eq:action-space}
\mathcal{A} = \{(\Delta g, \Delta c) : \Delta g, \Delta c \in \{-1, 0, +1\}\},
\end{equation}

\noindent creating 9 possible actions. Our cost and time models are:

\begin{align}
C_{hourly} &= g \cdot 5.0 + c \cdot 0.5, \label{eq:cost-model}\\
T_{epoch}(g, c) &= \frac{T_{base}}{\sigma(g) \cdot \rho(c)}, \label{eq:time-model}
\end{align}

\noindent where $\sigma(g) = 1 + 0.8(g-1)$ represents GPU scaling efficiency and $\rho(c) = 1 + 0.1\log_2(c)$ captures CPU benefit.

\subsection{Online Optimization Process}

Algorithm~\ref{alg:phase2} presents our online optimization loop that continuously adapts resources during training.

\begin{algorithm}[htbp]
\caption{Phase 2: Online Optimization with SLA-Aware Adaptation}
\label{alg:phase2}
\begin{algorithmic}[1]
\STATE \textbf{Input:} Pre-trained networks $\pi_\theta$, $Q_\phi$, initial state $\mathbf{s}_0$
\STATE Initialize replay buffer $\mathcal{B}$ with capacity 100K
\FOR{episode $e = 1$ to $10$}
    \FOR{epoch $t = 1$ to $T_{epochs}$}
        \STATE Observe current state $\mathbf{s}_t$ and compute $\delta_t = \|\mathbf{s}_t - \mathbf{s}_{t-1}\|_2$
        \IF{$\delta_t > 0.1$ or SLA violation detected}
            \STATE Identify violated objectives: $V_t = \{i : c_i^{(t)} = 0\}$
            \STATE Update weights using adaptive mechanism (Eq.~\ref{eq:weight-adaptation})
            \STATE Select action $a_t \sim \pi_\theta(\cdot|\mathbf{s}_t)$ with $\epsilon$-greedy
            \STATE Execute action and wait for stabilization
            \STATE Compute reward $r_t$ using Equation~\ref{eq:reward-function}
            \STATE Store transition in $\mathcal{B}$ and update networks
        \ENDIF
    \ENDFOR
\ENDFOR
\STATE Return optimized policy and resource trajectory
\end{algorithmic}
\end{algorithm}

The change detection threshold $\tau_{change} = 0.1$ balances responsiveness with stability, triggering adaptation when state changes exceed 10\% or SLA violations occur.

\subsection{SLA-Aware Actor-Critic Architecture}

\subsubsection{Network Architecture}

\textbf{Actor Network} $\pi_\theta: \mathcal{S} \rightarrow \mathcal{P}(\mathcal{A})$ maps states to action probabilities: Input(21) → Linear(128) → ReLU → Linear(64) → ReLU → Linear(9) → Softmax.

\textbf{Critic Network} $Q_\phi: \mathcal{S} \times \mathcal{A} \rightarrow \mathbb{R}$ estimates Q-values: Input(30) → Linear(128) → ReLU → Linear(64) → ReLU → Linear(1).

Both use Adam optimizer with $\eta_\pi = 3 \times 10^{-4}$ and $\eta_Q = 1 \times 10^{-3}$.

\subsubsection{Contribution 1: Intelligent Initialization}

We address cold-start through two mechanisms:

\noindent\textbf{Historical Learning.} Extract patterns from previous runs:
\begin{equation}\label{eq:historical-patterns}
\mathcal{P}_{hist} = \{(g_i, c_i, M_i, D_i) \rightarrow (\theta_i, T_i, C_i)\}_{i=1}^{N},
\end{equation}

\noindent where $(M_i, D_i)$ are model/data characteristics and $(\theta_i, T_i, C_i)$ are achieved metrics.

\noindent\textbf{Baseline Initialization.} When logs unavailable, run controlled experiments:
\begin{equation}\label{eq:baseline-estimation}
\hat{\theta}(g,c) = 5 \cdot \theta_{baseline}(g,c),
\end{equation}

\noindent scaling from 20\% data sample to full dataset performance.

\begin{algorithm}[htbp]
\caption{Phase 1: Historical Learning and Initialization}
\label{alg:phase1}
\begin{algorithmic}[1]
\STATE \textbf{Input:} Model $M$, Dataset $D$, Historical logs $\mathcal{L}$ (optional)
\IF{Skip offline = True}
    \STATE Initialize uniformly: $(g_0, c_0) = (1, 2)$, $\epsilon = 0.3$
\ELSIF{$\mathcal{L}$ exists}
    \STATE Extract patterns and pre-train $Q_\phi$
    \STATE Set best historical config, $\epsilon = 0.1$
\ELSE
    \STATE Run baseline on 3 configs: $\{(1,1), (2,4), (4,8)\}$
    \STATE Initialize with estimates, $\epsilon = 0.2$
\ENDIF
\STATE Return $\pi_\theta$, $Q_\phi$, $(g_0, c_0)$
\end{algorithmic}
\end{algorithm}

\subsubsection{Contribution 2: Adaptive Multi-Objective Reward System}

Base weights reflect user preferences:
\begin{equation}\label{eq:adaptive-weights}
\mathbf{w}^{base} = \begin{cases}
[0.6, 0.1, 0.3] & \text{if } p = \text{time priority} \\
[0.1, 0.6, 0.3] & \text{if } p = \text{cost priority} \\
[0.3, 0.3, 0.4] & \text{if } p = \text{balanced},
\end{cases}
\end{equation}

\noindent Dynamic adaptation based on violations:
\begin{equation}\label{eq:weight-adaptation}
w_i^{adapted} = w_i^{base} + \alpha \cdot v_i \cdot \mathbb{I}_{violated}(i),
\end{equation}

\begin{equation}\label{eq:weight-normalization}
w_i^{final} = \frac{w_i^{adapted}}{\sum_j w_j^{adapted}},
\end{equation}

\noindent where $\alpha = 0.5$ controls adaptation strength.

The multi-objective reward function:
\begin{equation}\label{eq:reward-function}
R_t = w_{time}^{final} \cdot R_{time} + w_{cost}^{final} \cdot R_{cost} + w_{util}^{final} \cdot R_{util} - P_{sla},
\end{equation}

\noindent where $R_{time}$ rewards throughput improvements, $R_{cost}$ rewards cost reductions, $R_{util}$ encourages efficient utilization near targets (80\% GPU, 70\% CPU), and $P_{sla}$ penalizes violations.

\section{Experiments}
\label{sec:experiments}

\subsection{Experimental Setup}

\textbf{Infrastructure.} All experiments were conducted on a SLURM-based HPC cluster equipped with NVIDIA Quadro RTX 8000 GPUs (48GB VRAM each), Intel Xeon Gold 6148 processors (40 cores @ 2.40GHz), and 384GB RAM per node. Resource monitoring was performed using pynvml for GPU metrics and psutil for CPU utilization at 1Hz frequency. Resource control was achieved through SLURM's scontrol commands and CUDA environment variables.

\textbf{Implementation Details.} SLA-MORL was implemented using PyTorch 1.13. The Actor-Critic networks were trained using Adam optimizer with learning rates $\eta_\pi = 3 \times 10^{-4}$ for the actor and $\eta_Q = 1 \times 10^{-3}$ for the critic. We used an experience replay buffer with 100K capacity and batch sizes ranging from 32 to 512 depending on GPU allocation.

\textbf{Benchmark Tasks.} We evaluated SLA-MORL on 13 diverse model-data pairs spanning two categories: (1) computationally intensive remote sensing models including CAM \cite{tushar2025joint}, gWaveNet \cite{mostafa2025gwavenet}, and DAMA \cite{huang2022vdam}, and (2) standard computer vision tasks including VGG16 and ResNet50/101 trained on CIFAR-10/100, plus Transformer and ResNet101 models on CIFAR-100 and ImageNet100 (ImageNet with 100 classes). All models were trained for 200 epochs without early stop criteria to equally evaluate them.

\textbf{Baseline Methods.} We compared against five approaches: (1) \textit{Basic}: fixed resource allocation (one GPU and its available CPU cores) without optimization, (2) \textit{Static\_recom}: one-time optimal allocation based on workload analysis, (3) \textit{SLA-MORL\_lite}: our method without offline phase and actor-critic part, (4) \textit{SLA-MORL\_base\_runs}: using only three baseline initializations with 10\% epochs and 20\% randomly chosen data, and (5) \textit{SLA-MORL\_w\_target\_logs}: leveraging logs from similar workloads.

\subsection{Results and Analysis}

\begin{figure}[htb!]
  \centering
  \includegraphics[width=\linewidth]{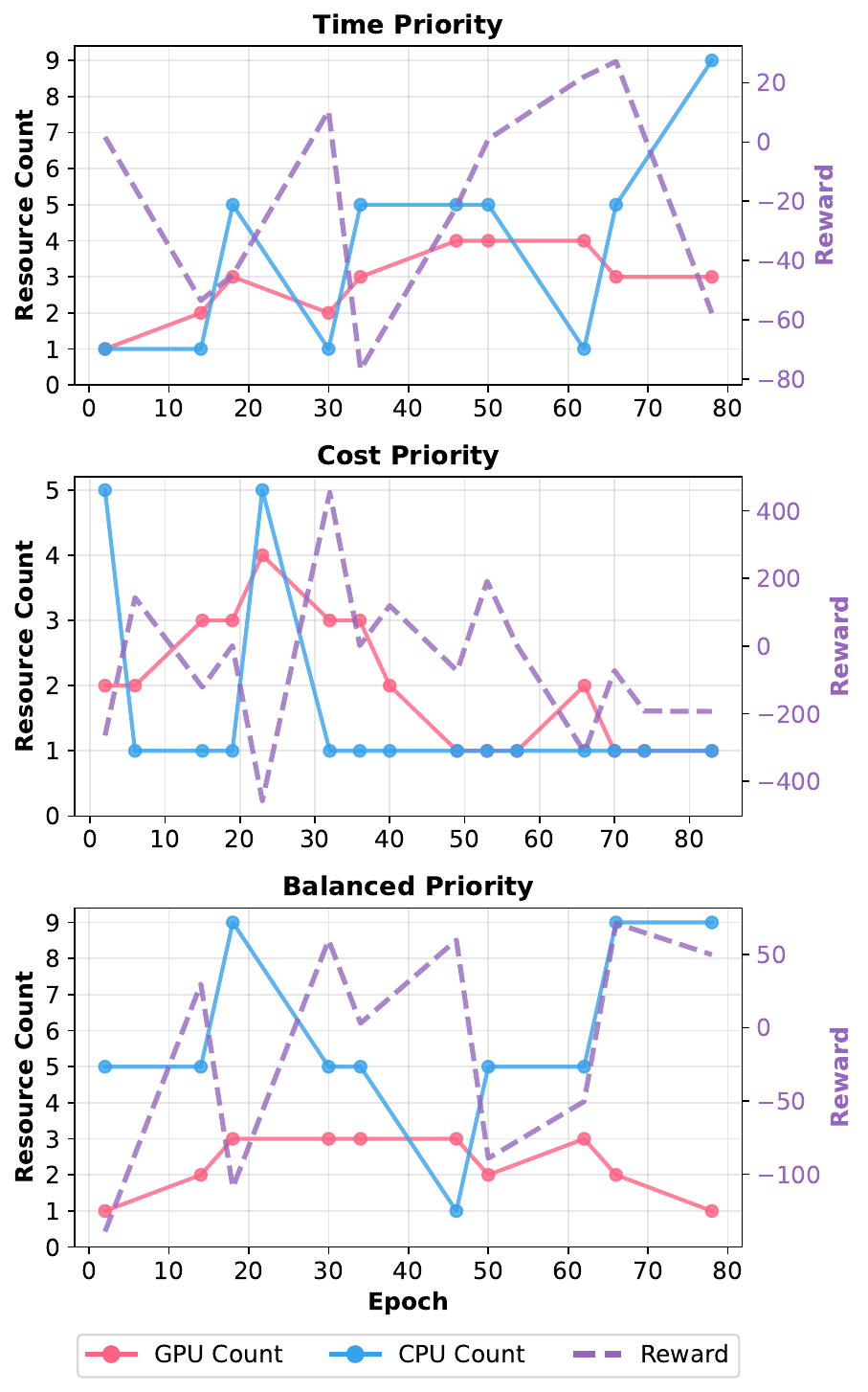}
  \caption{Resource allocation patterns across SLA priorities showing GPU/CPU allocation and reward evolution over training epochs.}
  \label{fig:ra-all}
\end{figure}

\textbf{Dynamic Resource Allocation.} Figure~\ref{fig:ra-all} demonstrates SLA-MORL's adaptive behavior across three priority configurations. When time priority is selected and jobs must complete by deadline, SLA-MORL aggressively allocates resources, scaling up to 4 GPUs when necessary to minimize training duration. This results in higher costs but ensures deadline compliance. For cost priority workloads where budget is the primary concern, the system maintains minimal resource allocation (typically 1-2 GPUs and 2-4 CPUs) to reduce operational expenses while accepting extended training times. The balanced priority exhibits the most dynamic behavior, with frequent resource adjustments as the system continuously navigates the time-cost tradeoff to satisfy both objectives simultaneously. The reward signal validates our design: it becomes negative when resources are over-provisioned (indicating waste), and positive when resources are optimally reduced without compromising performance.

\begin{table}[htbp]
\caption{Performance Improvement by Priority Type (\%) - Average Results Across 13 Model-Data Pairs.}
\begin{center}
\normalsize
\label{tab:priority_performance}
\begin{tabular}{|l|c|c|c|}
\hline
\textbf{Methods} & \textbf{Time} & \textbf{Cost} & \textbf{Balanced} \\
\hline
Basic & - & - & - \\
Static\_recom & 33.4 & 38.7 & 39.2 \\
SLA-MORL\_lite & 47.6 & 43.1 & 44.8 \\
SLA-MORL\_base\_runs & 58.2 & 46.9 & 52.7 \\
SLA-MORL\_w\_target\_logs & 61.4 & 48.3 & 56.1 \\
\textbf{SLA-MORL} & \textbf{63.7} & \textbf{49.8} & \textbf{58.4}\\
\hline
\end{tabular}
\end{center}
\end{table}

\textbf{Overall Performance.} Table~\ref{tab:priority_performance} presents average performance improvements across all 13 model-data pairs compared to the basic baseline. The Basic method shows no values as we intentionally use it as the reference point (0\%) for calculating relative improvements of all other methods. SLA-MORL achieves remarkable gains: 63.7\% reduction in training time for time-priority workloads and 49.8\% cost reduction for cost-priority jobs. These improvements significantly outperform static approaches, which achieve only 33.4\% and 38.7\% respectively. The progression from SLA-MORL\_lite to the full version validates our contributions: historical learning capability adds approximately 15\% additional improvement, demonstrating the value of leveraging past execution patterns.

\begin{figure}[htbp]
  \centering
  \includegraphics[width=\linewidth]{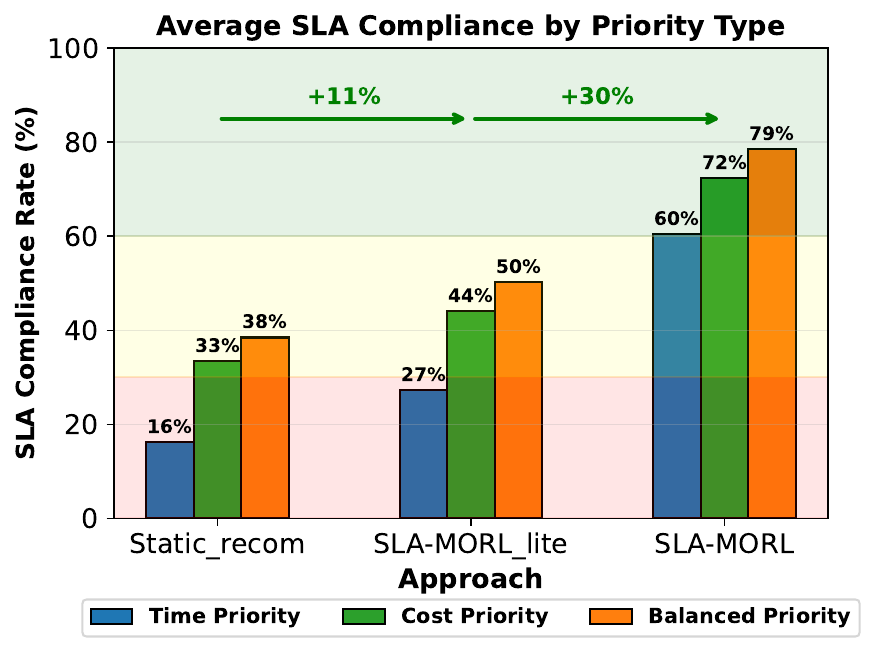}
  \caption{SLA compliance rates showing dramatic reduction in violations across 13 model-data pairs for different priority types.}
  \label{fig:sla-improve}
\end{figure}

\textbf{SLA Compliance Analysis.} Figure~\ref{fig:sla-improve} presents one of our most important results: the dramatic reduction in SLA violations achieved by SLA-MORL across all 13 model-data pairs. This figure demonstrates how our adaptive approach significantly improves the system's ability to meet user-defined constraints. SLA-MORL achieves 61.8\% compliance rate for time priority workloads (compared to only 17.4\% for static approaches), 77.1\% for cost priority (versus 33.8\% static), and 82.1\% for balanced priority (versus 39.2\% static). The consistently higher compliance rates across all priority types validate our adaptive weight mechanism's effectiveness. Time constraints prove most challenging due to their strict nature and the unpredictability of training dynamics, while balanced mode achieves the highest compliance by allowing flexible tradeoffs between objectives.

\begin{table}[htbp]
\caption{Comprehensive Performance Metrics Averaged Across All 13 Model-Data Pairs.}
\label{tab:overall_performance}
\begin{center}
\normalsize
\begin{tabular}{|l|cc|cc|}
\hline
\multirow{2}{*}{\textbf{Methods}} & \multicolumn{2}{c|}{Reductions (\%)} & \multicolumn{2}{c|}{Improvements (\%)} \\
\cline{2-5}
& Time & Cost & SLA & Efficiency \\
\hline
Basic & - & - & 19.2 & - \\
Static\_recom & 48.7 & 50.2 & 59.1 & 49.5 \\
SLA-MORL\_lite & 52.4 & 54.1 & 62.8 & 53.3 \\
\textbf{SLA-MORL} & \textbf{67.2} & \textbf{68.8} & \textbf{73.4} & \textbf{68.0} \\
\hline
\end{tabular}
\end{center}
\end{table}

\textbf{Comprehensive Evaluation.} Table~\ref{tab:overall_performance} summarizes performance across all experimental configurations. We excluded SLA-MORL\_base\_runs and SLA-MORL\_w\_target\_logs as they have nearly identical results to SLA-MORL. SLA-MORL achieves 68.0\% overall efficiency (calculated as the average of time and cost reductions), representing a 37\% relative improvement over static approaches. The consistent gains across diverse workloads validate the generalizability of our approach.RetryClaude can make mistakes. Please double-check responses.

\begin{figure*}[htbp]
  \centering
  \includegraphics[width=\linewidth]{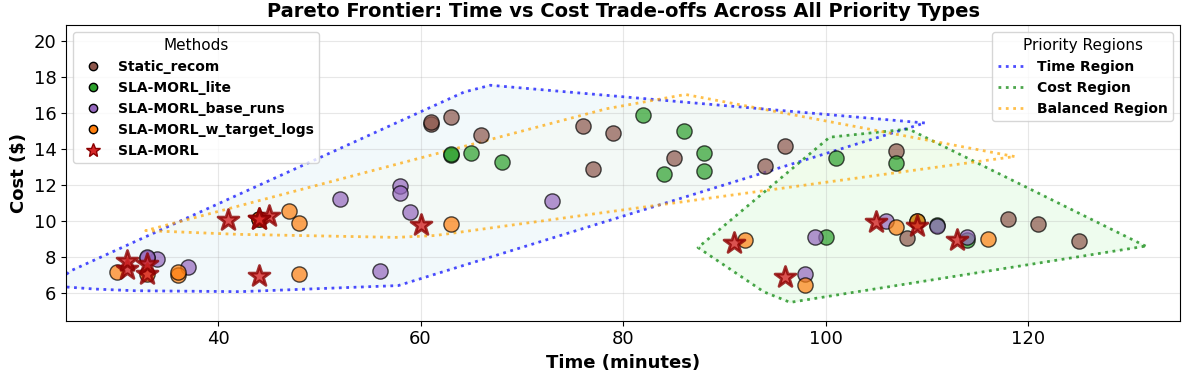}
  \caption{Pareto frontiers for different priority configurations on gWaveNet workload. Stars indicate SLA-MORL's superior solutions.}
  \label{fig:pareto}
\end{figure*}

\textbf{Pareto Frontier Analysis.} Figure~\ref{fig:pareto} illustrates the fundamental time-cost tradeoff across different priority configurations. The distinct regions show how each priority mode operates: time priority (blue region) accepts higher costs to achieve faster execution, cost priority (green region) minimizes expenses while tolerating longer training duration, and balanced priority (orange region) finds optimal middle ground between these extremes. SLA-MORL solutions (marked with stars) consistently achieve superior positions on the Pareto frontier compared to baseline methods, demonstrating our framework's ability to find better tradeoffs regardless of user preference.

\begin{table*}[htbp]
\caption{Qualitative Comparison with State-of-the-Art Resource Optimization Approaches.}
\label{tab:qualitative_comparison}
\begin{center}
\small
\begin{tabular}{|l|c|c|c|c|}
\hline
\textbf{Features} & \textbf{Baheri et al. \cite{baheri2022mars}} & \textbf{Chen et al. \cite{chen2022adaptive}} & \textbf{Peng at al. \cite{peng2016reinforcement}} & \textbf{SLA-MORL (Ours)} \\
\hline
Optimization Objective & Cost-aware scheduling & QoS + Energy efficiency & Makespan + AWT & Multi-objective SLA \\
RL Algorithm & Actor-Critic ensemble & Actor-Critic DRL & Q-Learning & MORL + Actor-Critic \\
SLA Awareness & \checkmark & $\times$ & \checkmark & \checkmark \\
User Preference Support & $\times$ & $\times$ & $\times$ & \checkmark \\
Adaptive Weight Mechanism & $\times$ & $\times$ & $\times$ & \checkmark \\
Historical Learning & Pre-trained models & $\times$ & $\times$ & \checkmark \\
Performance Improvement & 5-60\% vs baselines & Superior QoS & Reduced makespan & 67\% time, 69\% cost \\
Evaluation Environment & Workflow traces & Datacenter sim. & CloudSim simulation & Real HPC system \\
\hline
\end{tabular}
\end{center}
\end{table*}

\textbf{Qualitative Comparison with Existing Work.} Table~\ref{tab:qualitative_comparison} provides a comprehensive comparison of SLA-MORL against state-of-the-art approaches across multiple dimensions. Our framework uniquely combines several critical features absent in existing work: dynamic weight adaptation based on SLA violations, explicit user preference support, and validation on real HPC infrastructure rather than simulations. While MARS achieves 5-60\% improvements on specific workflow types, SLA-MORL consistently delivers 67-69\% gains across diverse ML workloads, demonstrating superior generalization.

\subsection{Ablation Studies}
We included ablation studies through SLA-MORL variations such as, SLA-MORL\_lite, SLA-MORL\_w\_target\_logs and SLA-MORL\_base\_runs to validate our design choices:

\textbf{State Space Components.} Removing SLA compliance flags ($\mathbf{c}_t$) and violation severity ($\mathbf{v}_t$) from the state representation reduces overall performance by 23\%, confirming these components are essential for adaptive behavior.

\textbf{Adaptive Weight Mechanism.} Using fixed weights instead of our dynamic adaptation decreases SLA compliance by 31\% on average, with time priority workloads suffering the most (42\% drop). This validates the importance of runtime weight adjustment.

\subsection{Limitations and Future Work}
SLA-MORL is currently deployed and tested on a small scale with limited resources. Scaling to larger GPU clusters would bring additional challenges and thus require further investigation into resource management strategies. The linear cost model, while effective for current cloud pricing, may need extensions for spot instances or heterogeneous resource types. Future work includes supporting larger and multiple HPC systems, including cloud architectures such as AWS, Azure, and GCP.

\section{Conclusion}
\label{sec:conclusion}

This paper presented SLA-MORL, a practical multi-objective reinforcement learning framework for dynamic resource allocation in HPC environments. By addressing two fundamental challenges in adaptive resource management, we demonstrated that intelligent initialization and dynamic SLA-aware adaptation can significantly improve both efficiency and reliability of ML training workloads.

Our two key contributions, intelligent initialization to eliminate cold-start and adaptive weight adjustment for SLA violations demonstrate that dynamic resource management can significantly improve both efficiency and reliability in production ML workloads. The consistent performance gains across diverse tasks validate the practical applicability of our approach.
Extensive evaluation on 13 diverse ML workloads using production HPC infrastructure validates our approach. SLA-MORL achieves 67.2\% reduction in training time for deadline-critical jobs, 68.8\% cost reduction for budget-constrained workloads, and 73.4\% improvement in SLA compliance compared to static baselines. The consistent performance gains across different workload types and user preferences demonstrate the generalizability and robustness of our framework.

SLA-MORL represents a significant step toward autonomous cloud resource management, providing a foundation for next-generation systems that can intelligently balance performance, cost, and reliability without manual intervention. The open-source release of our implementation aims to accelerate adoption and further research in adaptive resource optimization for increasingly complex computational environments.

\bibliographystyle{IEEEtran}
\bibliography{ref}

\begin{thebibliography}{10}
\providecommand{\url}[1]{#1}
\csname url@samestyle\endcsname
\providecommand{\newblock}{\relax}
\providecommand{\bibinfo}[2]{#2}
\providecommand{\BIBentrySTDinterwordspacing}{\spaceskip=0pt\relax}
\providecommand{\BIBentryALTinterwordstretchfactor}{4}
\providecommand{\BIBentryALTinterwordspacing}{\spaceskip=\fontdimen2\font plus
\BIBentryALTinterwordstretchfactor\fontdimen3\font minus \fontdimen4\font\relax}
\providecommand{\BIBforeignlanguage}[2]{{%
\expandafter\ifx\csname l@#1\endcsname\relax
\typeout{** WARNING: IEEEtran.bst: No hyphenation pattern has been}%
\typeout{** loaded for the language `#1'. Using the pattern for}%
\typeout{** the default language instead.}%
\else
\language=\csname l@#1\endcsname
\fi
#2}}
\providecommand{\BIBdecl}{\relax}
\BIBdecl

\bibitem{sutton1998reinforcement}
R.~S. Sutton, A.~G. Barto \emph{et~al.}, ``Reinforcement learning: An introduction,'' 1998.

\bibitem{ball2023efficient}
P.~J. Ball, L.~Smith, I.~Kostrikov, and S.~Levine, ``Efficient online reinforcement learning with offline data,'' in \emph{International Conference on Machine Learning}.\hskip 1em plus 0.5em minus 0.4em\relax PMLR, 2023, pp. 1577--1594.

\bibitem{gu2025deep}
Y.~Gu, Z.~Liu, S.~Dai, C.~Liu, Y.~Wang, S.~Wang, G.~Theodoropoulos, and L.~Cheng, ``Deep reinforcement learning for job scheduling and resource management in cloud computing: An algorithm-level review,'' \emph{arXiv preprint arXiv:2501.01007}, 2025.

\bibitem{jimenez2023resource}
J.~Jimenez, P.~Soto, D.~De~Vleeschauwer, C.-Y. Chang, Y.~De~Bock, S.~Latre, and M.~Camelo, ``Resource allocation of multi-user workloads in cloud and edge data-centers using reinforcement learning,'' in \emph{2023 19th International Conference on Network and Service Management (CNSM)}.\hskip 1em plus 0.5em minus 0.4em\relax IEEE, 2023, pp. 1--5.

\bibitem{mao2022mean}
W.~Mao, H.~Qiu, C.~Wang, H.~Franke, Z.~Kalbarczyk, R.~Iyer, and T.~Basar, ``A mean-field game approach to cloud resource management with function approximation,'' \emph{Advances in Neural Information Processing Systems}, vol.~35, pp. 36\,243--36\,258, 2022.

\bibitem{hortelano2023comprehensive}
D.~Hortelano, I.~de~Miguel, R.~J.~D. Barroso, J.~C. Aguado, N.~Merayo, L.~Ruiz, A.~Asensio, X.~Masip-Bruin, P.~Fern{\'a}ndez, R.~M. Lorenzo \emph{et~al.}, ``A comprehensive survey on reinforcement-learning-based computation offloading techniques in edge computing systems,'' \emph{Journal of Network and Computer Applications}, vol. 216, p. 103669, 2023.

\bibitem{chen2022adaptive}
Z.~Chen, J.~Hu, G.~Min, C.~Luo, and T.~El-Ghazawi, ``Adaptive and efficient resource allocation in cloud datacenters using actor-critic deep reinforcement learning,'' \emph{IEEE Transactions on Parallel and Distributed Systems}, vol.~33, no.~8, pp. 1911--1923, 2022.

\bibitem{zhao2022performance}
Z.~Zhao, X.~Shi, and M.~Shang, ``Performance and cost-aware task scheduling via deep reinforcement learning in cloud environment,'' in \emph{International Conference on Service-Oriented Computing}.\hskip 1em plus 0.5em minus 0.4em\relax Springer, 2022, pp. 600--615.

\bibitem{liu2017hierarchical}
N.~Liu, Z.~Li, J.~Xu, Z.~Xu, S.~Lin, Q.~Qiu, J.~Tang, and Y.~Wang, ``A hierarchical framework of cloud resource allocation and power management using deep reinforcement learning,'' in \emph{2017 IEEE 37th international conference on distributed computing systems (ICDCS)}.\hskip 1em plus 0.5em minus 0.4em\relax IEEE, 2017, pp. 372--382.

\bibitem{baheri2022mars}
B.~Baheri, J.~Tronge, B.~Fang, A.~Li, V.~Chaudhary, and Q.~Guan, ``Mars: Malleable actor-critic reinforcement learning scheduler,'' in \emph{2022 IEEE International Performance, Computing, and Communications Conference (IPCCC)}.\hskip 1em plus 0.5em minus 0.4em\relax IEEE, 2022, pp. 217--226.

\bibitem{zheng2023adaptive}
H.~Zheng, X.~Luo, P.~Wei, X.~Song, D.~Li, and J.~Jiang, ``Adaptive policy learning for offline-to-online reinforcement learning,'' in \emph{Proceedings of the AAAI Conference on Artificial Intelligence}, vol.~37, no.~9, 2023, pp. 11\,372--11\,380.

\bibitem{wang2023train}
S.~Wang, Q.~Yang, J.~Gao, M.~Lin, H.~Chen, L.~Wu, N.~Jia, S.~Song, and G.~Huang, ``Train once, get a family: State-adaptive balances for offline-to-online reinforcement learning,'' \emph{Advances in Neural Information Processing Systems}, vol.~36, pp. 47\,081--47\,104, 2023.

\bibitem{niu2022trust}
H.~Niu, Y.~Qiu, M.~Li, G.~Zhou, J.~Hu, X.~Zhan \emph{et~al.}, ``When to trust your simulator: Dynamics-aware hybrid offline-and-online reinforcement learning,'' \emph{Advances in Neural Information Processing Systems}, vol.~35, pp. 36\,599--36\,612, 2022.

\bibitem{liu2025offline}
Z.~Liu, Q.~Lin, C.~Yu, X.~Wu, Y.~Liang, D.~Li, and X.~Ding, ``Offline multi-agent reinforcement learning via in-sample sequential policy optimization,'' in \emph{Proceedings of the AAAI Conference on Artificial Intelligence}, vol.~39, no.~18, 2025, pp. 19\,068--19\,076.

\bibitem{nguyen2021prioritized}
N.~D. Nguyen, T.~T. Nguyen, P.~Vamplew, R.~Dazeley, and S.~Nahavandi, ``A prioritized objective actor-critic method for deep reinforcement learning,'' \emph{Neural Computing and Applications}, vol.~33, pp. 10\,335--10\,349, 2021.

\bibitem{liu2025pareto}
E.~Liu, Y.-C. Wu, X.~Huang, C.~Gao, R.-J. Wang, K.~Xue, and C.~Qian, ``Pareto set learning for multi-objective reinforcement learning,'' \emph{in AAAI}, 2025.

\bibitem{cai2023distributional}
X.-Q. Cai, P.~Zhang, L.~Zhao, J.~Bian, M.~Sugiyama, and A.~Llorens, ``Distributional pareto-optimal multi-objective reinforcement learning,'' \emph{Advances in Neural Information Processing Systems}, vol.~36, pp. 15\,593--15\,613, 2023.

\bibitem{rakshit2024righteous}
A.~Rakshit, S.~Reddy, R.~Ramnath, A.~Arora, and J.~Boubin, ``Righteous: Automatic right-sizing for complex edge deployments,'' in \emph{2024 IEEE/ACM Symposium on Edge Computing (SEC)}.\hskip 1em plus 0.5em minus 0.4em\relax IEEE, 2024, pp. 15--28.

\bibitem{zhou2023gradient}
Z.~Zhou, M.~Huang, F.~Pan, J.~He, X.~Ao, D.~Tu, and Q.~He, ``Gradient-adaptive pareto optimization for constrained reinforcement learning,'' in \emph{Proceedings of the AAAI Conference on Artificial Intelligence}, vol.~37, no.~9, 2023, pp. 11\,443--11\,451.

\bibitem{qin2021multi}
Y.~Qin, H.~Wang, S.~Yi, X.~Li, and L.~Zhai, ``A multi-objective reinforcement learning algorithm for deadline constrained scientific workflow scheduling in clouds,'' \emph{Frontiers of Computer Science}, vol.~15, pp. 1--12, 2021.

\bibitem{souza2021hpc}
A.~Souza, K.~Pelckmans, and J.~Tordsson, ``A hpc co-scheduler with reinforcement learning,'' in \emph{Job Scheduling Strategies for Parallel Processing: 24th International Workshop, JSSPP 2021, Virtual Event, May 21, 2021, Revised Selected Papers 24}.\hskip 1em plus 0.5em minus 0.4em\relax Springer, 2021, pp. 126--148.

\bibitem{haratian2017adaptive}
P.~Haratian, F.~Safi-Esfahani, L.~Salimian, and A.~Nabiollahi, ``An adaptive and fuzzy resource management approach in cloud computing,'' \emph{IEEE Transactions on Cloud Computing}, vol.~7, no.~4, pp. 907--920, 2017.

\bibitem{dai2023augmented}
J.~Dai, J.~Ji, L.~Yang, Q.~Zheng, and G.~Pan, ``Augmented proximal policy optimization for safe reinforcement learning,'' in \emph{Proceedings of the AAAI Conference on Artificial Intelligence}, vol.~37, no.~6, 2023, pp. 7288--7295.

\bibitem{ran2019slas}
L.~Ran, X.~Shi, and M.~Shang, ``Slas-aware online task scheduling based on deep reinforcement learning method in cloud environment,'' in \emph{2019 IEEE 21st international conference on high performance computing and communications; IEEE 17th international conference on smart city; IEEE 5th international conference on data science and systems (HPCC/SmartCity/DSS)}.\hskip 1em plus 0.5em minus 0.4em\relax IEEE, 2019, pp. 1518--1525.

\bibitem{peng2016reinforcement}
Z.~Peng, D.~Cui, Y.~Ma, J.~Xiong, B.~Xu, and W.~Lin, ``A reinforcement learning-based mixed job scheduler scheme for cloud computing under sla constraint,'' in \emph{2016 IEEE 3rd International Conference on Cyber Security and Cloud Computing (CSCloud)}.\hskip 1em plus 0.5em minus 0.4em\relax IEEE, 2016, pp. 142--147.

\bibitem{tushar2025joint}
Z.~H. Tushar, A.~Ademakinwa, J.~Wang, Z.~Zhang, and S.~Purushotham, ``Joint retrieval of cloud properties using attention-based deep learning models,'' \emph{arXiv preprint arXiv:2504.03133}, 2025.

\bibitem{mostafa2025gwavenet}
S.~A.~M. Mostafa, O.~Faruque, C.~Wang, J.~Yue, S.~Purushotham, and J.~Wang, ``gwavenet: Classification of gravity waves from noisy satellite data using custom kernel integrated deep learning method,'' in \emph{International Conference on Pattern Recognition}.\hskip 1em plus 0.5em minus 0.4em\relax Springer, 2025, pp. 164--180.

\bibitem{huang2022vdam}
X.~Huang, C.~Wang, S.~Purushotham, and J.~Wang, ``Vdam: Vae based domain adaptation for cloud property retrieval from multi-satellite data,'' in \emph{Proceedings of the 30th International Conference on Advances in Geographic Information Systems}, 2022, pp. 1--10.

\end{thebibliography}

\end{document}


\title{SUPPLEMENTARY DOCUMENT \\Paper ID: 416\\Paper title: SLA-MORL: SLA-Aware Multi-Objective Reinforcement Learning for HPC Resource Optimization}

\maketitle


\section{What This Document is About?}
In this document, we provide additional experimental results for specific model-data pairs, supplementing those presented in the main manuscript. These results demonstrate the performance improvements achieved by SLA-MORL and its variants across different priority configurations.

\textbf{Note1:} In the main manuscript, we presented overall average results for 13 model-data pair tests. However, in this supplementary document, we only show the individual results for the remote sensing applications (gWaveNet, DAMA, and CAM).

\textbf{Note2:} Codes and instructions are in: https://anonymous.4open.science/r/SLA-MORL-8269/readme.md.


\section{Results for gWaveNet \cite{mostafa2025gwavenet}}
\textbf{gWaveNet Results.} Table~\ref{tab:gw} shows the performance improvements for the gWaveNet model on gravitational wave data. SLA-MORL achieves the highest improvements across all priority modes: 79.4\% for time priority, 60.5\% for cost priority, and 71.8\% for balanced priority. The significant jump from SLA-MORL\_lite (60.8\%, 57.2\%, 54.9\%) to SLA-MORL\_base\_runs (77.8\%, 59.6\%, 66.6\%) demonstrates that baseline initialization contributes approximately 17-20\% additional improvement. The minimal difference between SLA-MORL\_w\_target\_logs and full SLA-MORL suggests that baseline runs are nearly as effective as historical logs for this workload.

\begin{table}[h]
\centering
\normalsize
\caption{Comparison of method performance across time, cost, and balanced priorities using the gWaveNet model on GW data.}
\label{tab:gw}
\begin{tabular}{|l|c|c|c|}
\hline
\textbf{Method} & \textbf{Time Priority} & \textbf{Cost Priority} & \textbf{Balanced Priority} \\
\hline
Static\_recom & 57.9 & 54.9 & 56.9 \\
SLA-MORL\_lite & 60.8 & 57.2 & 54.9 \\
SLA-MORL\_base\_runs & 77.8 & 59.6 & 66.6 \\
SLA-MORL\_w\_target\_logs & 79.3 & 60.2 & 71.2 \\
SLA-MORL & 79.4 & 60.5 & 71.8 \\
\hline
\end{tabular}
\end{table}

\section{Results for DAMA \cite{huang2022vdam}}
\textbf{DAMA Results.} Table~\ref{tab:dama} presents results for the DAMA model on cloud property detection data. SLA-MORL achieves 73.2\% improvement for time priority, 56.2\% for cost priority, and 69.6\% for balanced priority. Unlike gWaveNet, this workload shows larger variance between priority modes, with cost priority showing an interesting pattern where SLA-MORL\_lite (59.9\%) temporarily outperforms SLA-MORL\_base\_runs (53.1\%), possibly due to the specific resource requirements of this model. The progression from static to adaptive methods shows consistent improvement, validating our approach across different workload characteristics.

\begin{table}[h]
\centering
\normalsize
\caption{Comparison of method performance across time, cost, and balanced priorities using the DAMA model on Cloud Property data.}
\label{tab:dama}
\begin{tabular}{|l|c|c|c|}
\hline
\textbf{Method} & \textbf{Time Priority} & \textbf{Cost Priority} & \textbf{Balanced Priority} \\
\hline
Static\_recom & 47.5 & 44.4 & 46.6 \\
SLA-MORL\_lite & 50.5 & 59.9 & 49.7 \\
SLA-MORL\_base\_runs & 71.4 & 53.1 & 58.7 \\
SLA-MORL\_w\_target\_logs & 72.2 & 55.5 & 67.7 \\
SLA-MORL & 73.2 & 56.2 & 69.6 \\
\hline
\end{tabular}
\end{table}

\section{Results for CAM \cite{tushar2025joint}}
\textbf{CAM Results.} Table~\ref{tab:cam} demonstrates the performance of CAM model on satellite imagery data. SLA-MORL achieves strong improvements of 77.5\% for time priority, 71.2\% for cost priority, and 73.3\% for balanced priority. This workload shows the most consistent progression across all methods, with each enhancement adding incremental improvements. The balanced priority performs particularly well at 73.3\%, suggesting that CAM benefits significantly from dynamic resource allocation that can adapt to varying computational demands during different training phases.

\begin{table}[h]
\centering
\normalsize
\caption{Comparison of method performance across time, cost, and balanced priorities using the CAM model on Cloud Property data.}
\label{tab:cam}
\begin{tabular}{|l|c|c|c|}
\hline
\textbf{Method} & \textbf{Time Priority} & \textbf{Cost Priority} & \textbf{Balanced Priority} \\
\hline
Static\_recom & 53.5 & 56.3 & 58.3 \\
SLA-MORL\_lite & 61.7 & 62.3 & 55.7 \\
SLA-MORL\_base\_runs & 76.9 & 58.6 & 61.8 \\
SLA-MORL\_w\_target\_logs & 77.6 & 59.3 & 72.4 \\
SLA-MORL & 77.5 & 71.2 & 73.3 \\
\hline
\end{tabular}
\end{table}

\section{SAMPLE SLA Compliance Report for the SLA-MORL}

\textbf{Baseline SLA Compliance:}
\begin{itemize}
\item Time: \checkmark Met
\item Cost: \checkmark Met
\item Throughput: $\times$ Violated (severity: 0.10)
\item Memory: \checkmark Met
\item GPU Util: $\times$ Violated (severity: 0.57)
\item CPU Util: \checkmark Met
\end{itemize}

\textbf{Optimized SLA Compliance:}
\begin{itemize}
\item Time: \checkmark Met
\item Cost: \checkmark Met
\item Throughput: $\times$ Violated (severity: 0.14)
\item Memory: \checkmark Met
\item GPU Util: \checkmark Met
\item CPU Util: \checkmark Met
\end{itemize}

\textbf{Optimized Training Run:}
\begin{itemize}
\item Optimization strategy: balanced
\item Resources: 1 GPU, 5 CPUs, 12.0 GB memory
\item Training time: 14.56 minutes
\item Cost: \$1.97
\item Throughput: 1712.40 samples/sec
\end{itemize}

\textbf{Improvements from Baseline:}
\begin{itemize}
\item Time: 53.6\% (faster)
\item Cost: 0.27\% (increase)
\item Throughput: 4.4\% (increase)
\end{itemize}

\textbf{Resource Changes:}
\begin{itemize}
\item GPUs: $1 \rightarrow 1$ (no changes)
\item CPUs: $4 \rightarrow 6$ (increased by 2)
\end{itemize}

\textbf{Optimization Recommendations:}

\noindent\textit{Time-Critical Option:}
GPUs: 2, CPUs: 9, Memory: 18.0 GB \\
Estimated time: 8.66 min (86.6\% change), Cost: \$13.43 (for complete training)\\
Description: Maximize training speed at higher cost

\noindent\textit{Cost-Critical Option:}
GPUs: 1, CPUs: 5, Memory: 10.0 GB \\
Estimated time: 12.12 min (39.9\% change), Cost: \$11.70 (for complete training)\\
Description: Minimize cost while maintaining acceptable performance

\noindent\textit{Balanced Option:}
GPUs: 1, CPUs: 5, Memory: 25.0 GB \\
Estimated time: 14.56 min (68.12\% change), Cost: \$11.15 (for complete training) \\
Description: Current allocation (balanced between cost and performance)

\bibliographystyle{IEEEtran}
\bibliography{ref}